\documentclass[smallextended]{svjour3}
\smartqed 
\PassOptionsToPackage{numbers, sort&compress}{natbib}
\usepackage{natbib}

\usepackage[margin=1in]{geometry}
\usepackage{times}  
\usepackage{helvet}  
\usepackage{courier}  
\usepackage{graphicx}  

\usepackage{color}
\usepackage{xspace}
\usepackage{amsmath,amssymb}
\usepackage{graphicx}

\usepackage[T1]{fontenc}    
\usepackage[utf8]{inputenc} 
\usepackage{booktabs}       
\usepackage{amsfonts}       
\usepackage{nicefrac}       
\usepackage{microtype}      

\newcommand{\ie}{i.e.\xspace}

\newcommand{\apriori}{\emph{a priori}\xspace}

\newcommand{\mat}[1]{\mathbf{#1}}
\renewcommand{\vec}[1]{ \mathbf{#1} } 
\newcommand{\vecS}[1]{\boldsymbol{ #1 }  } 

\newcommand{\F}{\mat{F}}
\newcommand{\G}{\mat{G}}
\newcommand{\I}{\mat{I}}
\newcommand{\K}{\mat{K}}

\newcommand{\X}{\mat{X}}
\newcommand{\W}{\mat{W}}
\newcommand{\Y}{\mat{Y}}
\newcommand{\Z}{\mat{Z}}

\newcommand{\calL}{\mathcal{L}}

\newcommand{\calO}{\mathcal{O}}

\newcommand{\setR}{\mathbb{R}}

\newcommand{\kron}{\otimes}

\newcommand{\expectation}[2]{ \mathbb{E}_{#1}{\left[#2\right]} }

\newcommand{\covariance}[2]{\left\langle #1, #2 \right\rangle}
\newcommand{\Normal}{\mathcal{N}}

\newcommand{\trace}{\mbox{ \rm tr }}
\renewcommand{\det}[1]{\left\lvert#1\right\rvert}
\newcommand{\defeq}{\stackrel{\text{\tiny def}}{=}}

\newcommand{\bigO}{\calO}

\newcommand{\dataset}{\mathcal{D}}

\newcommand{\adelautumn}{\name{adel-autm}}
\newcommand{\adelsummer}{\name{adel-summ}}
\newcommand{\sydautumn}{\name{syd-autm}}
\newcommand{\melws}{\name{wind}} 

\newcommand{\ggp}{\name{ggp}}

\newcommand{\gptext}{{\sc gp}\xspace}

\newcommand{\modg}{{\sc m}o{\sc g}\xspace}
\newcommand{\mog}{{\sc m}o{\sc g}\xspace}

\newcommand{\name}[1]{{\sc #1}\xspace}
\newcommand{\igp}{\name{igp}}
\newcommand{\mtg}{\name{mtg}}

\newcommand{\gprn}{\name{gprn}}

\newcommand{\lcm}{\name{lcm}}

\newcommand{\adam}{\name{adam}}

\newcommand{\elbotext}{\name{elbo}}

\newcommand{\elbo}{\calL_{\text{elbo}}}
\newcommand{\df}{\text{d}\f}

\newcommand{\ellterm}{\calL_{\text{ell}}}

\newcommand{\enterm}{\calL_{\text{ent}}}
\newcommand{\crossterm}{\calL_{\text{cross}}}
\newcommand{\elltermhat}{\widehat{\calL}_{\text{ell}}}

\newcommand{\nlpd}{\name{nlpd}}

\newcommand{\rmse}{\name{rmse}}

\newcommand{\fvar}{\name{f-var}}
\newcommand{\mrank}{\name{m-rank}}

\newcommand{\gp}{\mathcal{GP}}
\newcommand{\meanfun}{\mu}
\newcommand{\kernel}{\kappa}

\newcommand{\hyperparam}{\vecS{\theta}}
\newcommand{\likeparam}{\vecS{\phi}}

\newcommand{\pcond}[2]{p(#1  | #2 )}

\newcommand{\bs}{\boldsymbol}
\newcommand{\llambda}{\bs{\lambda}}

\newcommand{\s}{S} 
\newcommand{\n}{N} 
\newcommand{\m}{M} 
\newcommand{\p}{P} 
\newcommand{\q}{Q} 
\renewcommand{\d}{D} 
\renewcommand{\k}{K} 
\newcommand{\dimh}{H} 

\newcommand{\vecy}{\vec{y}} 
\newcommand{\y}{\Y} 
\newcommand{\yn}{\vecy_{(n)}} 
\newcommand{\yni}{\vecy_{(n) i}} 

\newcommand{\xn}{\vec{x}_{(n)}}
\newcommand{\f}{\F} 
\newcommand{\vecf}{\vec{f}} 
\newcommand{\fn}{\vecf_{(n)}} 
\newcommand{\fnnodot}{\vecf_{(n)}} 
\newcommand{\fj}{\vecf_{j}} 

\renewcommand{\u}{\vec{u}} 
\newcommand{\x}{\vec{x}}
\newcommand{\xprime}{\x^\prime}

\newcommand{\z}{\mat{Z}}
\renewcommand{\X}{\mat{X}}
\newcommand{\vecz}{\vec{z}}

\newcommand{\lags}{\boldsymbol{l}} 
\newcommand{\h}{\vec{h}} 
\newcommand{\bR}{R} 
\newcommand{\br}{r} 
\newcommand{\Qr}{Q_r} 
\newcommand{\Qg}{Q_g} 

\newcommand{\fr}{\f_{\br}} 
\newcommand{\frn}{\vecf_{\br(n)}} 
\newcommand{\Wn}{\mat{W}_{(n)}} 
\newcommand{\wni}{\vec{w}_{(n)i}} 

\newcommand{\g}{\mat{G}}
\newcommand{\vecg}{\vec{g}}
\newcommand{\gj}{\vec{g}_j}  
\newcommand{\gjprime}{\vec{g}_{j'}}  
\newcommand{\gn}{\vec{g}_{(n)}} 

\newcommand{\Krxxhh}{\K_{f f}^\br} 
\newcommand{\Krxx}{\K_{\x\x}^\br} %

\newcommand{\Krhh}{\K_{\h\h}^\br} 
\newcommand{\Krhhinv}{(\K_{\h\h}^\br)^{-1}} 
\newcommand{\Krzz}{\K_{\vecz\vecz}^\br} 
\newcommand{\Krzzinv}{(\K_{\vecz\vecz}^\br)^{-1}} 
\newcommand{\Krxz}{\K_{\x\vecz}^\br} 
\newcommand{\Krzzhh}{\K_{u u}^\br} 
\newcommand{\Krzzhhinv}{(\K_{u u}^\br)^{-1}} 
\newcommand{\Krxzh}{\K_{f u}^\br} 
\newcommand{\Krzxh}{\K_{u f}^\br} 

\newcommand{\ur}{\vec{u}_{ \br} } 
\newcommand{\Zr}{\z_\br} 
\newcommand{\priormeanr}{\tilde{\vecS{\mu}}_\br} 
\newcommand{\priorcovr}{\widetilde{\K}_\br}
\newcommand{\Ar}{\mat{A}_\br}

\newcommand{\Hrmat}{\mat{H}_\br}
\newcommand{\priorcovrn}{\widetilde{\K}_{\br (n)}}
\newcommand{\Arn}{\mat{A}_{\br (n)}}

\newcommand{\krxh}{\kernel_{\br}(f_j(\x), f_{j'}(\xprime) )}

\newcommand{\krxx}{\kernel_{\br}(\x, \x')}
\newcommand{\krxxn}{\kernel_{\br}(\x_{(n)}, \x_{(n)}}
\newcommand{\krxzn}{\kernel_{\br}(\x_{(n)},\Z_r)}
\newcommand{\krzxn}{\kernel_{\br}(\Z_r, \x_{(n)})}

\newcommand{\krhh}{\kernel_{\br}( \h_j, \h_{j'} )}
\newcommand{\ksub}[1]{\kernel_{#1} }

\newcommand{\kprop}{\pi_k}
\newcommand{\qukr}{q_k(\ur | \llambda_{k \br})}
\newcommand{\varparamkr}{\llambda_{k \br}}

\newcommand{\xstar}{\x_{\star}}
\newcommand{\fstar}{\vecf_{\star}}

\newcommand{\ystar}{\vecy_{\star}}

\renewcommand{\K}{\mat{K}}

\newcommand{\Kxx}{\K_{\x\x}^j}

\newcommand{\vectheta}{\vecS{\theta}}
\newcommand{\kj}{\kernel_{j}(\x, \x'; \vectheta_j )}

\newcommand{\postmean}[1]{\vec{m}_{#1}}
\newcommand{\postcov}[1]{\mat{S}_{#1}}

\newcommand{\plike}{\pcond{\Y}{\f, \likeparam}}
\newcommand{\logpliken}{\log \pcond{\yn}{\fn, \likeparam}}
\newcommand{\qjoint}{q(\f, \u | \llambda)}
\newcommand{\qu}{q(\u | \llambda)}
  
\newcommand{\qf}{q(\f | \llambda)}
  
\newcommand{\qkfstar}{q_k(\fstar | \llambda_k)}  
\newcommand{\qknf}{q_{k(n)}(\fn | \llambda_k )}
\newcommand{\qnf}{q_{(n)}(\fnnodot | \llambda)}
\newcommand{\qfstar}{q(\fstar | \llambda)} 

\newcommand{\qfmean}[1]{\vec{b}_{#1}}        
\newcommand{\qfcov}[1]{\mat{\Sigma}_{#1}}        

\newcommand{\iid}{iid\xspace}

\newcommand{\eq}{Eq.\xspace}
\newcommand{\eqs}{Eqs.\xspace}

\newcommand{\rbf}{\name{rbf}}

\title{Grouped Gaussian Processes for Solar Power Prediction}

\author{Astrid Dahl \and Edwin V.~Bonilla}

 \institute {Astrid Dahl
 	\at School of Computer Science and Engineering, University of New South Wales, Sydney
 	\\\email{astridmdahl@gmail.com}
 	\and 
 	Edwin V.~Bonilla
 	\at Data61, Sydney
 	 }

\begin{document}
\maketitle

\begin{abstract}
We consider multi-task regression models where the observations are assumed to be a linear combination of several latent node functions and weight functions, which are both drawn from Gaussian process priors. Driven by the problem of developing scalable methods for forecasting distributed solar and other renewable power generation, we propose coupled priors over groups of (node or weight) processes to exploit spatial dependence between functions. We estimate forecast models for solar power at multiple distributed sites and ground wind speed at multiple proximate weather stations. Our results show that our approach maintains or improves point-prediction accuracy relative to competing solar benchmarks and improves over wind forecast benchmark models on all measures. Our approach consistently dominates the equivalent model without coupled priors, achieving faster gains in forecast accuracy. At the same time our approach provides better quantification of predictive uncertainties.
\end{abstract}

\section{Introduction}
\label{sec:intro}
The problem of forecasting local solar output in the short term is of significant interest for the purpose of distributed grid control and household energy management \citep{voyant-et-al-solar-2017,Widen-et-al-solar-2015}.  Variation in output is driven by two principal factors: diurnal cyclical effects (variation due to sun angle and distance) and variability due to weather effects, both inducing spatially-related dependence between proximate sites. 
In general, correlations across sites depend on many particulars relating to system configuration, local environment and so on. As such we wish to exploit spatial dependencies (and potentially other site-specific covariates) between sites in a flexible manner. More importantly, inherent to this application is the need for modeling uncertainty in a flexible and principled way \citep{antonanzas2016review}.

Gaussian process (\gptext) models are a flexible nonparametric Bayesian approach that can be applied to various problems such as regression and classification \citep{rasmussen-williams-book} and have been extended to numerous multivariate and multi-task problems  including spatial and spatio-temporal contexts \citep{cressie-wikle-book-2011}. Multi-task \gptext methods have been developed along several lines \citep[see e.g.][for a review]{Alvarez:2012:KVF:2344402.2344403}. Of relevance here are various mixing approaches that combine multiple latent univariate Gaussian processes via linear or nonlinear mixing to predict multiple related tasks \citep{wilson-et-al-icml-12}. The challenge in multi-task cases is maintaining scalability of the approach. To this end, both scalable inference methods and model constraints have been employed \citep{alvarez2010efficient,GPflow2017,krauth-et-al-uai-2017}. In particular, latent Gaussian processes are generally constrained to be statistically independent \citep{wilson-et-al-icml-12,krauth-et-al-uai-2017}.

In this paper we consider the case where the statistical independence constraint is relaxed for subsets of latent functions. We build on the scalable generic inference method of \citet{krauth-et-al-uai-2017} to extend the model of \citet{wilson-et-al-icml-12} and allow nonzero covariance between arbitrary subsets, or `groups' of latent functions. The grouping structure is flexible and can be tailored to applications of interest, and additional features can potentially be incorporated to govern input-dependent covariance across functions. By adopting separable kernel specifications, we maintain scalability of the approach whilst capturing latent dependence structures.

With this new multi-task \gptext model, we consider the specific challenge of forecasting power output of multiple, distributed residential solar sites. We apply our approach to capture spatial covariance between sites by explicitly incorporating spatial dependency between latent functions and test our method on three datasets comprised of solar output at distributed household sites in Australia. 

For many of the same reasons, short term wind power forecasting represents significant challenges yet is critical to emerging energy technologies \citep{Widen-et-al-solar-2015}. Output variability is driven by wind speed, which (as for solar) is driven by multiple interacting environmental factors giving rise to spatial dependencies \apriori. To demonstrate the broader applicability of the model, we also illustrate our approach on a wind speed dataset comprised of ground wind speed at distributed weather stations in Australia.

Our results show that, for solar models, introducing spatial covariance over groups of latent functions maintains or improves point-prediction forecast accuracy relative to competing benchmark methods and at the same time provides better quantification of predictive uncertainties. Further, wind forecast accuracy and uncertainty is improved on all measures by the introduction of spatial covariance.  Timed experiments show that the new model dominates the equivalent model without spatial dependencies, achieving similar or superior forecast accuracy in a shorter time.
\section{Related work}
Gaussian Processes have been considered in the multi-task setting via a number of approaches. Several methods linearly combine multiple univariate \gptext models via coefficients that may be parameters \citep[latent factor models as in][]{teh-et-al-aistats-05}; linear coregional models \citep[\lcm;][]{goovaerts-book}, or themselves input dependent \citep{wilson-et-al-icml-12}.

Most mixing approaches focus on methods to combine multiple underlying independent latent functions. Recent developments in inference for multi-task \gptext models have improved scalability of mixing approaches, building upon the variational framework of \citet{titsias-aistats-2009}. \citet{nguyen-bonilla-nips-2014} develop a generic variational inference method that allows efficient optimization for multi-task models with arbitrary likelihoods, while the sparse, variational framework of \citep{hensman-et-al-aistats-2015,GPflow2017} supported significant gains in scalability of multi-task \gptext models. \citet{dezfouli-bonilla-nips-2015} extend the approach in \citep{nguyen-bonilla-nips-2014} to the sparse variational context, exploiting inducing points to improve scalability of the inference method using a general mixture-of-Gaussian sparse, variational posterior. More recently, the approach of \citep{dezfouli-bonilla-nips-2015} was extended to integrate optimization that exploits leave-one-out objective learning in addition to the sparse, variational lower bound \citep{bonilla-et-al-arxiv-2016,krauth-et-al-uai-2017}.

Other multi-task \gptext approaches allow task-specific predictions through use of task-specific features or `free-form' cross-task covariances \citep{bonilla-et-al-nips-08}, and more recently priors placed over cluster allocations allowing cluster-specific covariances \citep{hensman-fast14,gardner-et-al-arxiv-2018}. Combination via convolutions has also been developed and extended to sparse, variational settings \citep{alvarez-lawrence-nips-08,alvarez2011computationally}. 

Coupling between $\q$ node (but not weight) latent functions directly is considered  by \citep{remes-pmlr-2017}, who build upon the Gaussian process regression network (\gprn) framework of \citep{wilson-et-al-icml-12}. 
The authors propose a rich, Generalized Wishart-Gibbs kernel that characterizes covariance for latent functions. The fully-coupled kernel is internally parameterized rather than utilizing feature-dependent cross-function covariance. The approach makes use of variational inference to approximate the model. Unlike our method, however, the natural disadvantage of such an approach is that it presents significant computational challenges in terms of scalability to a large number of observations and tasks. This is primarily due to the need for variational inference that requires batch optimization with $\bigO((\n \q)^3)$  complexity, rendering it infeasible for larger scale applications. In fact, only small experiments were carried out in \citep{remes-pmlr-2017} with $\n \q$ in the order of (approximately) $100$ to $500$, since the approach is primarily developed for small-data problems requiring a highly expressive latent covariance structure.

\subsection{Multi-task solar power forecasting}
A number of studies have confirmed that multi-task learning approaches can be useful for distributed solar irradiance or solar power forecasting, finding that cross-site information is relevant \citep{yang-solar-2013,yang-solarlasso-2015}.
Several studies build on the early work of \citep{sampson-solar-1992} and consider co-kriging methods for distributed solar irradiance forecasting or spatial prediction (notably \citep{yang-solar-2013,shinozake-solar-2016}). Other approaches include a range of linear statistical methods, shown to be competitive at shorter horizons, and neural network methods \citep{inman-solar-2013,Widen-et-al-solar-2015,voyant-et-al-solar-2017}. 
These approaches are generally constrained by data requirements, notably pre-flattening of data to remove diurnal cyclical trends, which requires knowledge of local system and environment variables. 
In the context of small scale, distributed residential sites, such information is often unavailable, motivating approaches that do not rely on rich data history or feature sets as are typically required by current approaches \citep{inman-solar-2013,Widen-et-al-solar-2015,voyant-et-al-solar-2017,antonanzas2016review,yang2018history}

In addition to kriging studies, \gptext models have been considered for short term solar forecasting \citep{bilionis-solar-2014,dahl-bonilla-2017}. Earlier approaches are generally constrained to small-data problems by poor scalability of exact \gptext models.
More recently, \citep{dahl-bonilla-2017} use scalable sparse, variational inference to apply multi-task Gaussian (\mtg) and linear coregional models (\lcm) to forecast solar output at multiple, distributed residential sites. Multi-task approaches are found to improve model performance in mixed weather conditions, less so in sunny conditions. The specifications adopted, however, did not show strong improvement in overall forecast accuracy relative to the naive, univariate site \gptext benchmarks, with the \lcm performing significantly worse than \mtg and individual models in that setting. Moreover, the \mtg presents scalability challenges since inducing inputs in the sparse, variational framework adopted are shared across all observations and tasks. 
\section{Multi-task Gaussian process models}
\label{sec:theory_gp}
A Gaussian process (\gptext, \citep{rasmussen-williams-book}) is formally defined as a distribution over functions such that  $f(\x) \sim \gp (\meanfun(\x), \kernel(\x, \xprime) )$ is a Gaussian process with mean function $\meanfun(\x)$ and covariance function $\kernel(\x, \xprime)$ iff any subset of function values $f(\x_1), f(\x_2), \ldots, f({\x_N})$ follows a Gaussian distribution with mean $\vecS{\mu}$ and covariance $\mat{K}$, which are obtained by evaluating the corresponding mean function and covariance function at the input points $\X = \{\x_1, \ldots, \x_N \}$.

Standard single-task \gptext regression assumes observations are \iid versions of the latent function values corrupted by Gaussian noise. In this case, posterior inference can be carried analytically \citep[][chap.~2]{rasmussen-williams-book}.  In this paper we consider the more general case of multiple outputs, which sometimes is referred in the literature to as multi-task \gptext regression. In other words, we are given data of  the form $\dataset = \{\X \in \setR^{\n \times \d},  \Y \in \setR^{\n \times \p} \}$ where each $\x_{(n)}$ in $\X$ is a $\d$-dimensional vector of input features and each
$\yn$ in $\Y$ is a $\p$-dimensional vector of task outputs. Furthermore, we are interested in the case of generally non-linear (non-Gaussian) likelihoods, for which there is no analytically tractable posterior.  
\subsection{Latent Gaussian process models with independent priors \label{sec:basic-mtl}}
Fortunately, advances in variational inference \citep{kingma2013auto,rezende2014stochastic} have allowed the development of efficient posterior inference methods with `black-box' likelihoods. In the case of models with \gptext priors, \citet{krauth-et-al-uai-2017} have extended these results to modeling multiple outputs under non-linear likelihoods and \emph{independent} \gptext priors over multiple latent functions. In short, under such a modeling framework,  correlations between the $\p$ outputs using $\q$ independent latent functions $\{f_j(\x)\}_{j=1}^\q$ each drawn from a zero-mean \gptext prior, \ie $f_j(x) \sim \gp(\vec{0}, \kj)$, can be encoded via the  likelihood. As we shall see in \S \ref{sec:groupgprn}, an example of this is when the independent \gptext{s} are linearly combined via a set of weights, which can be deterministic as in the semi-parametric latent factor model of \cite{teh-et-al-aistats-05} or stochastic and input-dependent as in the \gprn of \cite{wilson-et-al-icml-12}. 

Therefore, within the framework in \citep{krauth-et-al-uai-2017} the prior over the latent function values corresponding to the $\n$ observations along with the likelihood model is given by:
\begin{align}
\label{eq:generic-model} 
\pcond{\F}{\hyperparam} &= \prod_{j=1}^{\q} \pcond{\fj}{ \hyperparam_{j}} = \prod_{j=1}^{\q} \Normal (\fj ; \vec{0},  \Kxx ) \text{,} \\
\quad
\plike &= \prod_{n=1}^{\n} \pcond{\yn}{\fn, \likeparam} \text{,}
\end{align}
where $\F$ is the $\n \times \q$ matrix of latent function variables; $\fj$ is the $\n$-dimensional vector for  latent function $j$;  and $\vectheta_j$ the corresponding hyper-parameters; $\fn$ is the $\q$-dimensional vector of latent function values corresponding to observation $n$ and $\likeparam$ are the likelihood parameters.

\citet{krauth-et-al-uai-2017} exploit the structure of the model in \eq \ref{eq:generic-model} to develop a scalable  algorithm via variational inference. While the likelihood in this model is suitable for most unstructured machine-learning problems such as standard regression and classification, the prior can be too restrictive for problems where dependences across tasks can be incorporated  explicitly. In this paper, driven by the solar power prediction problem where spatial relatedness can be leveraged to improve predictions across different sites, we lift this statistical independence (across latent functions) constraint in the prior to propose a new multi-task model where some of the functions are \emph{coupled a priori}.
\section{Grouped priors for multi-task GP models}
\label{sec:group-priors}
To group latent functions \apriori, we can define arbitrarily chosen subsets of latent functions in $\f$, $\fr, \br = 1, \ldots, \bR$, 
where $\bR$ is the total number of groups. For each group the number of latent functions within is denoted $\Qr$, which we will also refer to as the group size, with $\sum_{\br=1}^{\bR} \Qr = \q$. Each group is comprised of latent functions $\fr = \{ \fj \}_{j \in \text{ group } \br} $ and covariance between latent functions $f_j$ and $f_{j^\prime}$ is nonzero iff the functions $f_j$ and $f_{j^\prime}$ belong to the same group $r$.
 
Hence, the prior on $\f$ can be expressed similarly to the generic prior defined in \eq \eqref{eq:generic-model}:
\begin{equation}
\label{eq:priorlnk}
\pcond{\F}{\hyperparam} = \prod_{\br=1}^{\bR} \pcond{\fr}{ \hyperparam_{r}} = \prod_{\br=1}^{\bR} \Normal (\fr ; \vec{0},  \Krxxhh ),
\end{equation}
where $\Krxxhh \in \setR^{\n\Qr \times \n\Qr}$ is the covariance matrix generated by the group kernel function $\krxh$, which evaluates the covariance of functions $f_j$ and $f_{j^\prime}$ at the locations $\x$ and $\xprime$, respectively.

This structure allows arbitrary grouping of latent functions depending on the application (in our case, groups are structured for distributed forecasting, discussed below). However we emphasize that our inference method allows grouping between \emph{any} latent functions in $\F$ and does not make any assumptions (beyond the standard \iid assumption) on the conditional likelihood. Hence, since our model allows dependences between latent functions \apriori, we refer to it as grouped Gaussian processes (\ggp). Although we develop a generic and efficient method for \ggp models in \S \ref{sec:inference}, our focus on this paper is on a particular class of flexible multi-task regression models referred in the literature to as  Gaussian process regression networks \cite[\gprn,][]{wilson-et-al-icml-12}. 
\subsection{Separable kernels}
Before describing how \gprn{s} fit into the framework of \citet{krauth-et-al-uai-2017} and how we generalize them to incorporate grouped priors, it is important to describe a simple yet efficient way of modeling correlations across groups. Once latent functions are coupled \apriori, scalability becomes an important consideration. Thus, although $\krxh$ is not constrained in terms of kernel choice, for the problem at hand we consider separable kernels of the form $\krxh = \krxx \krhh$. $\h$ are defined as $\dimh$-dimensioned feature vectors forming an additional feature matrix $\Hrmat \in \setR^{\Qr \times \dimh}$ that characterizes covariance across functions $\fj \in \fr$. We describe in \S \ref{sec:ggp-solar} below how $\Hrmat$ can be used to exploit spatial dependency between tasks.

This separable structure yields covariance matrices of the Kronecker form $ \Krxxhh = \Krhh \kron \Krxx$, where $\Krxx \in \setR^{\n \times \n}$ and $\Krhh \in \setR^{\Qr \times \Qr}$.
By adopting the Kronecker-structured prior covariance over functions within a group, we reduce the maximum dimension of required matrix inversions, allowing scalable inference.

\section{Grouped Gaussian process regression networks}
\label{sec:groupgprn}
\citet{wilson-et-al-icml-12} consider the case where the (noiseless) observations are a linear combination of  Gaussian processes, $\{ g_{\ell}(\x) \}$,   where the coefficients, $\{ w_{p \ell}(\x) \}$, are also input-dependent and drawn from Gaussian process priors. In other words, their conditional likelihood model for a single observation at input point $\x$ and task $p$ is given by:
\begin{align}
	y_p(\x) = \sum_{\ell=1}^{\Qg} w_{p \ell}(\x) g_\ell(\x) + \epsilon_p \text{, } p=1, \ldots \p \text{,}
\end{align}
where $\{w_{p \ell}(\x), g_\ell(\x) \}$ are drawn from independent \gptext priors and $\epsilon_p$ is a task-dependent Gaussian noise variable. This model is termed Gaussian process regression networks (\gprn{s}) by \citet{wilson-et-al-icml-12} and $\{w_{p \ell}\}$ and $\{g_{\ell}\}$ are referred to as weight functions and node functions, respectively. It is easy to see how \gprn{s} fit into the latent Gaussian process model formulation of \cite{krauth-et-al-uai-2017}, as described in \S \ref{sec:basic-mtl}. We simply make $\{w_{p\ell}\}$ and $\{g_\ell\}$ subsets of latent functions in $\{f_j\}_{j=1}^Q$ with  $\p\Qg$ weight functions and $\Qg$ node functions so that $\Qg(\p+1) = \q$. 

Given the observed data $\dataset$, for each latent process (over weights or node functions) we need to create as many latent variables as observations. Therefore, it is useful to conceptualize the weights as $\p\Qg$ latent variables of dimension $\n \times 1$ arranged into a tensor $\W \in \setR^{\p \times \Qg \times \n}$. Similarly, the node functions can be represented by $\Qg$ latent variables of dimension $\n \times 1$ arranged into  a tensor $\G \in \setR^{\Qg \times 1 \times \n}$.  Therefore, the conditional likelihood for input $\xn$ can be written in matrix form as
\begin{equation}
\pcond{\yn}{\fn, \likeparam} = \Normal(\yn; \Wn \gn, \mat{\Sigma}_y) \text{,}
\end{equation}
where the latent functions are given by node and weight functions, \ie $\fn = \{\Wn, \gn\}$;  the conditional likelihood parameters $\likeparam = \mat{\Sigma}_y$ and  $\mat{\Sigma}_y$ is a diagonal matrix. 
$\p$-dimensional outputs are constructed  at $\xn$ as the product of a $\p \times \Qg$ matrix of weight functions, $\Wn$, and $\Qg$-dimensional vector of node functions $\gn$.  
\subsection{Grouping structure}
\begin{figure}[t]
	\centering
	\includegraphics[width=0.7\linewidth]{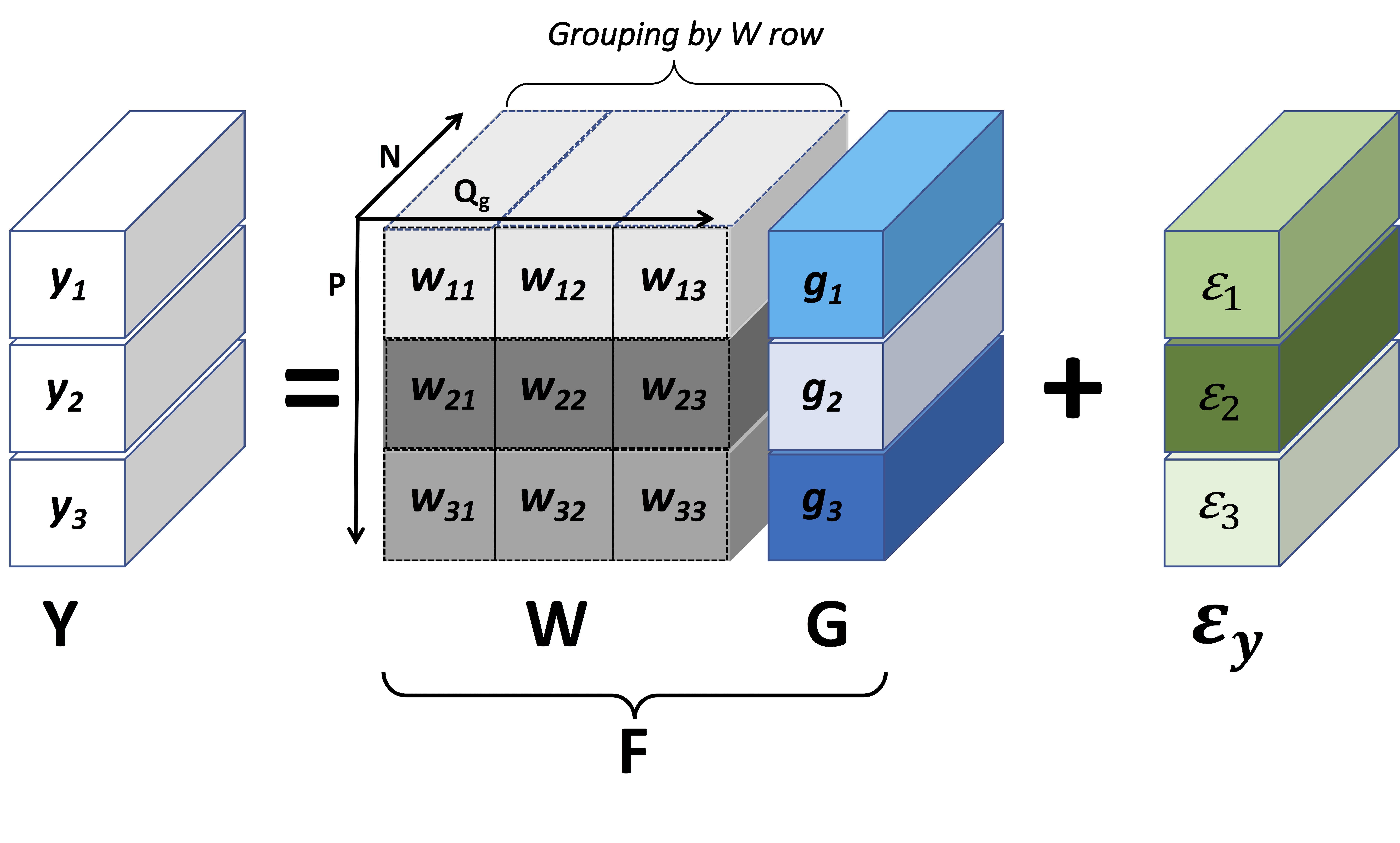}
	\caption{Gaussian process regression network model where $\y$ is a linear combination of node and weight latent functions comprising $\f$. In the grouped Gaussian process (\ggp) framework, latent functions may be grouped arbitrarily. A grouping scheme is illustrated where weight functions in $\W$ are grouped by rows (grouped functions are shown in the same shade) and given a fully-coupled prior, while node functions in $\G$ are independent. Here $\n$ is the number of observations per task; $\p$ is the number of tasks; and $\Qg$ is the group size.}
	\label{fig:ggp_block}
\end{figure}

Although our modeling and inference framework allows for arbitrary grouping structure, we consider a correlated prior over the rows of the weight functions for the grouped \gprn, and give details of the exact setting for the solar and wind applications in \S \ref{sec:ggp-solar}. Figure \ref{fig:ggp_block} illustrate our \ggp framework for the \gprn likelihood.

Naturally, the  greater flexibility of our approach comes at the expense of a high time-and-memory complexity, which poses significant challenges for posterior estimation. In the following section, we develop an efficient variational inference algorithm for our \ggp model  that is not much more computationally expensive than the original \gprn's. In fact, we show in our experiments in \S \ref{sec:timedexpts} that our inference method can converge faster than \gprn's while achieving similar or better performance. 
   
\section{Inference}
\label{sec:inference}
Our inference method is based on the generic inference approach for latent variable Gaussian process models set out by \citet{krauth-et-al-uai-2017}. This  is a sparse variational method that considers the case where latent functions are conditionally independent. We adapt that framework to the more general case where latent functions covary within groups, and for our case exploit the Kronecker structures noted at \S \ref{sec:groupgprn}. Since our inference method does not exploit any of the specifics of the \gprn likelihood, we consistently use the general grouped prior notation defined in \S \ref{sec:group-priors}.

Under the sparse method, the prior at \eqref{eq:priorlnk} is augmented with inducing variables, $\{ \ur \}_{\br = 1}^{\bR}$, drawn from the same \gptext priors as $\fr$ at new inducing points $\Zr$, where $\Zr \in \setR^{\m \times \d}$ lie in the same space as $\X \in \setR^{\n \times \d}$, $\m \ll \n$. Since $\ur$ are drawn from the same \gptext priors, inducing variables within a group $\br$ are similarly coupled via $\krxh$ evaluated at points $\Zr$.
The prior in \eq \eqref{eq:priorlnk} is thus replaced by
\begin{align}
\label{prior_sparse}
\pcond{\u}{ \hyperparam} &= \prod_{\br=1}^{\bR} \pcond{\ur}{ \hyperparam_{\br}} = \prod_{\br=1}^{\bR} \Normal (\ur ; \vec{0},  \Krzzhh )  \text{,}
\\
 \pcond{\f}{\u}  &= \prod_{\br=1}^{\bR} \Normal (\fr ; \priormeanr, \priorcovr ),  
 \end{align}
where  $\priormeanr = \Ar\ur$,  $\priorcovr = \Krxxhh - \Ar\Krzxh$  and $\Ar = \Krxzh\Krzzhhinv = \I_{Qr} \kron \Krxz\Krzzinv$. 
$\Krzzhh \in \setR^{\m\Qr \times \m\Qr}$ is the covariance matrix induced by $\krxh$ evaluated over $\Zr$, $\Hrmat$, yielding the structure $\Krzzhh = \Krhh \kron \Krzz$ and importantly the decomposition $\Krzzhhinv = \Krhhinv \kron \Krzzinv$.
We similarly define $\Krxzh$ and $\Krzxh$ (Table \ref{tab:modmats}).
\begin{table}[t]
	\centering
	\caption{A summary of the prior covariance matrix structures for a given group $r$ for scalable variational inference in the \ggp model.
		}
\noindent \begin{tabular}{@{}ll@{}ll@{}}
	 Notation & Specification &  Description\\
	\midrule
	$\Krxxhh$ & $\Krhh \kron \Krxx$  & Covariance between latent functions   \\
	$\Krzzhh$ & $\Krhh \kron \Krzz$ & Covariance between inducing variables \\
	$\Krxzh$ &  $\Krhh \kron \Krxz$ &  Covariance between latent functions and inducing variables\\
	$\priorcovr$ & $\Krxxhh - \Ar\Krzxh$ &  Covariance of conditional prior $p(\F | \u)$  \\
	$\Ar$ & $ \I_{Qr} \kron \Krxz\Krzzinv$ & Auxiliary matrix \\
	\bottomrule
\end{tabular}
\label{tab:modmats}%
\end{table}%
\subsection{Posterior estimation}
The (analytically intractable) joint posterior distribution of the latent functions and inducing variables under the prior and likelihood models in \eqs \eqref{prior_sparse} and \eqref{eq:generic-model} is approximated via variational inference \citep{jordan-etl-al-book-1998}. Specifically,
$
\pcond{\f,\u}{\y} = \pcond{\f}{\u, \y}\pcond{\u}{\y}  \approx \qjoint \defeq \pcond{\f}{\u}\qu.
$
The variational posterior $\qu$ is defined as a mixture of $\k$ Gaussians (\modg) with mixture proportions $\kprop$. We assume that $\qu$ also factorizes over blocks (and in the diagonal case over individual latent functions). The variational posterior is thus defined as
\begin{align}
\qu = \sum_{k=1}^{\k} \kprop \prod_{\br=1}^{\br} \qukr
\label{postvar}
\end{align}
where $\qukr = \Normal(\ur ; \postmean{k\br}, \postcov{k\br})$ and $\varparamkr = \{\postmean{k\br}, \postcov{k\br}, \kprop \}$.
We then estimate the model by maximizing the so-called evidence lower bound (\elbotext), which de-constructs to
$
\elbo (\llambda) \defeq
\enterm  (\llambda) + \crossterm (\llambda) + \ellterm (\llambda) \text{,}
$
which are the entropy, cross-entropy and expected log likelihood terms, respectively. 
The explicit expression required for $\elbo$  is a generalization of the results in \citet{krauth-et-al-uai-2017}. For the entropy term we have that (using Jensen's inequality): 
\begin{align}
\enterm(\llambda) &= \expectation{\qu}{\log \qu} 
 \nonumber \\
&\geq \sum_{k=1}^{\k} \kprop \sum_{l=1}^{\k}\Normal(\postmean{k}; \postmean{l}, \postcov{k} + \postcov{l}) ,
\label{eq_ent}
\end{align}
where $\postmean{k}$ is the vector $ \{ \postmean{k\br} \}_{\br=1}^{\bR} = \{ \postmean{kj} \}_{j=1}^{\q}$ and $\postcov{k}$ is the block diagonal matrix with diagonal elements $ \{ \postcov{k\br} \}_{\br=1}^{\bR}$ (and equivalent for $\postmean{l}$, $\postcov{l}$).
For the cross-entropy and the expected log likelihood terms: 
\begin{align}
\crossterm (\llambda) &= \expectation{\qu}{\log \pcond{\u}{ \hyperparam}} 
\nonumber \\
&= -\frac{1}{2} \sum_{k=1}^{\k} \kprop \sum_{\br=1}^{\bR} [\m_{\br} \log (2\pi) + \log \det{\Krzzhh} 
+ \postmean{k\br}'\Krzzhhinv \postmean{k\br} + \trace(\Krzzhhinv) \postcov{k\br} ]
\label{eq_cross} \\
& \nonumber \\ 
\ellterm (\llambda) &= \expectation{\qf}{\log \plike)} 
= \sum_{n=1}^{\n}\expectation{\qnf}{\logpliken}
\label{eq_ell}
\end{align}
where $\qf$ results from integration of the joint approximate posterior over inducing variables $\u$. Note that $\trace(\Krzzhhinv)$ factorizes as $ \trace(\Krzzinv) \trace(\Krhhinv)$  and $\ln \det{\Krzzhh}$ factorizes as $\m \ln \det{\Krzz} + \Qr \ln \det{\Krhh}$.
Given factorization of the  joint and variational posteriors over $k$ and $\br$ and standard conjugacy results, we have
\begin{align}
\qf &= \sum_{k=1}^{\k} \kprop \prod_{\br=1}^{\bR} \Normal (\qfmean{k\br}, \qfcov{k\br} ), 
\nonumber \\
\qfmean{k\br} &= \Ar\postmean{k\br}, \quad \text{and} \quad
 \qfcov{k\br} = \priorcovr + \Ar\postcov{k\br}\Ar' 
 \label{eq_qf}
\end{align}
The distribution $\qnf$ similarly factorizes as
\begin{align}
\qnf &= \sum_{k=1}^{\k} \kprop  \qknf 
\nonumber \\
&= \sum_{k=1}^{\k} \kprop \prod_{\br=1}^{\bR} \Normal ( \qfmean{k\br (n)}, \qfcov{k\br (n)} )  \text{.} 
\label{eq_qfn}
\end{align}
$\ellterm$ may be estimated by Monte Carlo, requiring sampling only from $\Qr$-dimensional multivariate Gaussians
$\Normal (\frn; \qfmean{k\br (n)}, \qfcov{k\br (n)} )$ where $\qfmean{k\br (n)}$ is the vector comprised of every $n$th element of $\qfmean{k\br}$, and $\qfcov{k\br (n)} $ is the (full) $\Qr \times \Qr$ matrix  comprised of $n$th diagonal elements of posterior covariance $\qfcov{k jj'}$ sub-matrices of $\qfcov{k\br} $.
Thus, we estimate
\begin{align}
\elltermhat = \frac{1}{\s} \sum_{n=1}^{\n} \sum_{k=1}^{\k} \kprop \sum_{s=1}^{\s} \ln \pcond{\yn}{\fn^{(k,s)}} \text{.}
\label{eq_ell_sampling}
\end{align}
Under the separable structure adopted, each mixture component covariance for $\frn$, $\qfcov{k\br (n)}$ can be seen to consist of structure arising from the grouped prior plus a term arising from the variational posterior:
\begin{align}
\qfcov{k\br (n)}  &= \priorcovrn + \Arn\postcov{k\br}\Arn^{\prime}, \quad \text{where}  \nonumber \\
\priorcovrn &= \Krhh \times \left[ \krxxn 
 - \krxzn \Krzz \krzxn   \right] \quad \text{and}  \nonumber \\
\Arn &= \left[ \I_{Qr} \kron \krxzn \Krzzinv \right]
\label{eq_postcov_n}
\end{align}
Thus cross-function covariance within a group will be either driven by the prior, where $\postcov{k\br}$ is diagonal, or more flexible in form where $\postcov{k\br}$ is non-diagonal.
\subsection{Prediction}
Prediction for a new point $\ystar$ given $\xstar$ is taken as the expectation over the general posterior distribution for the new point:
\begin{align}
\label{eq:pred-distro}
\pcond{\ystar}{\xstar} &=\int  \pcond{\ystar}{\fstar}\qfstar\df 
\nonumber \\
&= \sum_{k=1}^{\k} \kprop \int  \pcond{\ystar}{\fstar}\qkfstar \df_{\star} ,
\end{align}
where $\qkfstar$ is defined as for $\qknf$ in (\ref{eq_qfn}).
Given the explicit expression for the posterior distribution, the expectation in \eq \eqref{eq:pred-distro} is estimated by sampling:
\begin{align}
\expectation{\pcond{\ystar}{\xstar}}{\ystar} & \approx \frac{1}{\s} \sum_{s=1}^{\s} \W_{\star}^{s}\vecg_{\star}^{s} ,
\end{align}
where $\{ \W_{\star}^{s}\text{,}\vecg_{\star}^{s} \} = \fstar^{s}$ are samples from $\qkfstar$.
\subsection{Complexity}
Under the \ggp with a Kronecker-structured prior the time complexity per iteration changes slightly from the independent function case. For the same $\p, \Qg$ and $\m$, fewer $\m$-dimensioned inversions are required for \ggp versus \gprn, without any increase in maximum dimension under the Kronecker specification assuming $\m \geq \Qr$. This represents a substantial reduction in $\m$-dimensioned inversions, depending on the grouping scheme. 

The cost of calculating $\crossterm$ is dominated by the cost of inversions, being $\bigO \left(\sum_{\br=1}^{\bR} (\m^{3}+\Qr^{3}) \right)$ for the grouped case and $\bigO \left( \q\m^{3} \right)$ for the independent case. 
Under the diagonal posterior specification, $\postcov{k}$ in  \eq \eqref{eq_ent} reduces to the same form as the independent case of \citep{krauth-et-al-uai-2017}. Lastly, $\ellterm$ under the grouped structure requires sampling from low-dimensional $\Qr \times \Qr$ multivariate Gaussians with non-diagonal posterior covariance matrices, whereas this is avoided under the independent framework. However, the low dimensionality (number of tasks in our empirical evaluation) involved yields minimal additional cost.
\section{Grouped Gaussian processes for spatially dependent tasks \label{sec:ggp-solar}}
It is natural to consider a multi-task framework in a spatio-temporal setting such as distributed solar forecasting, where power output at solar sites in a region would \apriori be expected to covary over time and space. 
Given the expectation of spatially-driven covariance across sites, \ie tasks, we seek to exploit this structure to increase both efficiency and accuracy of multi-task forecasts. Our approach does this by incorporating explicit spatial dependencies between latent functions in the model.  

Latent functions in the general framework do not necessarily map to a particular task. The question therefore arises as to how to use spatial information relating to \emph{tasks} to structure covariance between \emph{latent functions}. 
We solve this by setting $\Qg = \p$ and grouping latent functions within rows of $\W$ \ie $\fj \in \W_{i:}, i=1,\dots \p$. We then define a feature matrix $\Hrmat$ that governs covariance across the $\p$ functions in each row (Figure \ref{fig:ggp_block}).
With $\Qg \geq \p$ it is possible to obtain a very general representation of the multi-task process with full mixing between tasks via $\g$, which now contains $\p$ node functions. This grouping structure allows parameters to vary across tasks, and at the same time, the coupled prior can act to regularize latent function predictions.

\paragraph{Model settings}
In our setting, we consider each latent process in $\g$ to be an independent \gptext, \ie, $\covariance{\gj}{\gjprime} = 0$ for $j \neq j^\prime$.  Furthermore, input features of $\gj, j= 1,\dots,\p$ are defined to be task features \ie features for $\gj$ relate to task $j$, specifically lagged-target values for $j$.

We define spatial features $\h_j = (latitude_{j}, longitude_{j})$ governing weightings applied to node functions. For a given task $i$,
$
\expectation{\pcond{\y}{\f, \likeparam}}{\yni} = \wni \gn
$
where $\wni$ denotes the $i$th row vector of $\Wn$. It can be seen that, in addition to depending on input features $\xn$, relative weights placed on node functions are now smoothed by spatial covariance imposed over the weights in $\wni$.
This allows site-by-site optimization of spatial decay in (cross-task) weights in addition to site-specific parameterization and features in $\wni$.
In total, this grouping structure yields $2\p$ groups: $\p$ groups of size $\p$  (corresponding to $\W$) and $\p$ groups of size $1$ (corresponding to $\g$).

Kernels and features for $\krxx$ and $\krhh$ are selected in line with previous studies relating to multi-task distributed forecasting \citep{inman-solar-2013,dahl-bonilla-2017}.
In particular, for our task of forecasting distributed solar output over time, for $\gj, j= 1,\dots,\p$, we define $\ksub{\gj}(\x_t,\x_s)= \ksub{\gj}(\lags_t,\lags_s)$ as a radial basis function kernel ($\kernel_{RBF}$) applied to a feature vector of recent lagged observed power at site $j$, \ie for site $j$ at time $t$, $\lags_{j,t} = (y_{j,t}, y_{j,t-1}, y_{j,t-2})$. 

For row-group $r$, we define a separable, multiplicative kernel structure as discussed above, \ie  
$\krxh = \krxx \krhh$. We set the kernel over the inputs as $\krxx = \kernel_{Per.}(t,s)\kernel_{RBF}(\lags_{rt},\lags_{rs})$, 
where $ \kernel_{Per.}(t,s)$ is a periodic kernel on a time index $t$ capturing diurnal cyclical trends in solar output. 

We adopt a compact kernel over functions, specifically a separable \rbf - Epanechnikov structure, \ie,  
$\krhh = \kernel_{RBF}(\h_j, \h_{j^\prime}) \kernel_{Ep.}(\h_j, \h_{j^\prime})$, $ j,j^\prime=1 \ldots \p$, where $\h_j = (latitude, longitude)$ for site $j$. By using a more flexible compact kernel, we aim to allow beneficial shared learning across tasks while reducing negative transfer by allowing cross-function weights to reduce to zero at an optimal distance.
\section{Experiments \label{sec:expts}}
We evaluate our approach on forecasting problems for distributed residential solar installations and wind speed measured at proximate weather stations. 
\subsection{Solar forecasting}
The task for solar is to forecast power production fifteen minutes ahead at multiple distributed sites in a region. 
Data consist of five minute average power readings from groups of proximate sites in the Adelaide and Sydney regions in Australia. We present results for three datasets: ten Adelaide sites (\adelautumn) and twelve Sydney sites (\sydautumn), both over 60 days during Autumn 2016, and ten Adelaide sites (\adelsummer) over 60 days in Spring-Summer 2016. We train all models on 36 days of data, and test forecast accuracy for 24 subsequent days (days are defined as 7 am to 7pm). In all, for each site, we have 5000 datapoints for training and 3636 datapoints for testing. 
\begin{figure}[htbp]
	\centering
	\includegraphics[width=0.7\linewidth]{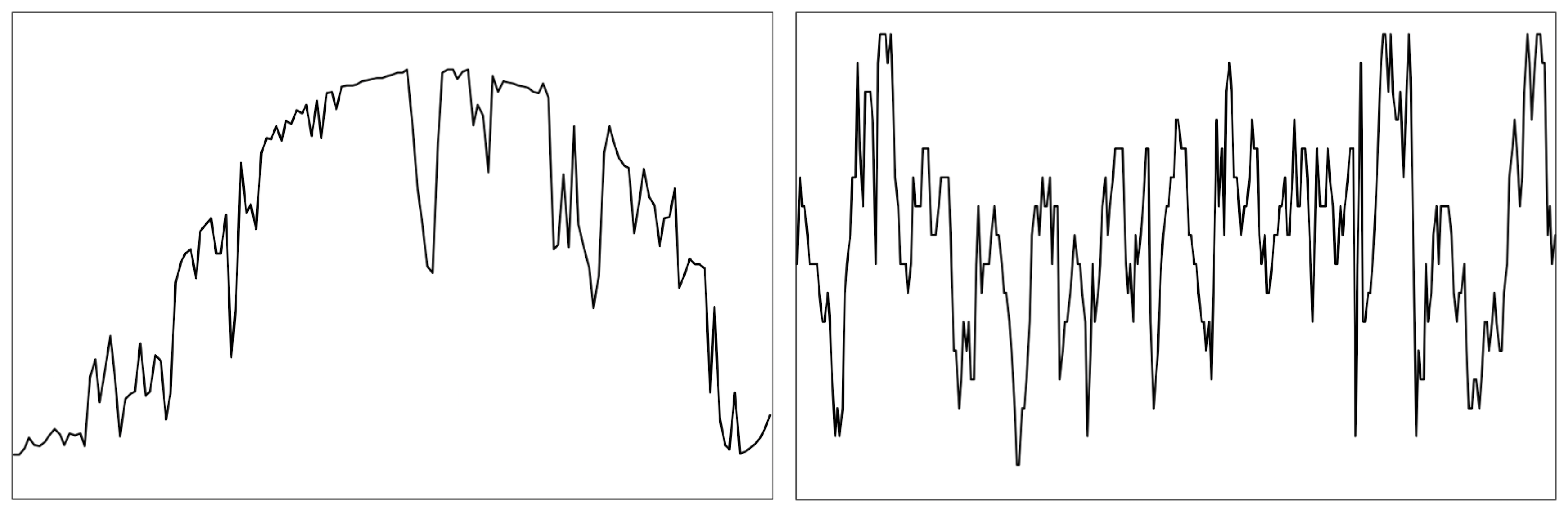}
	\caption{Example data for normalized solar power (\adelsummer - left hand side) and normalized wind speed (\melws - right hand side). Wind data exhibit strong noise relative to Summer time solar data.}
	\label{fig:data_example}
\end{figure}

Datasets have varying spatial dispersions. \adelautumn (\adelsummer) sites are spread over an approximately 30 by 40 (20 by 20) kilometer area, while \sydautumn sites are evenly dispersed over an approximately 15 by 20 kilometer  area.

\subsubsection{Benchmark models}
We compare forecast performance of our \ggp method  to the fully independent \gprn and several other benchmark models. We estimate (1) separate independent \gptext forecast models for each site  (\igp), (2) pooled multi-task models with task-specific (spatial) features for sites (\mtg), and (3) multi-task linear coregional models (\lcm). The final benchmark model (4) is the \gprn with independent latent functions \citep{wilson-et-al-icml-12}. 

These models can be expressed in terms of the general latent function framework with differing values of $\p$, $\q$ and $\bR$,  and different likelihood functions. As discussed in \S \ref{sec:groupgprn}, where latent functions are independent, group size is equal to 1 and $\bR = \q$. Key model constants are presented at Table \ref{tab:modconsts}. 

Both \igp and \mtg models have a standard, single task Gaussian likelihood functions, while the \lcm model is comprised of $\p$ node functions mapped to outputs via a $\p \times \Qg$  matrix of deterministic weights, 
\ie $\pcond{\yn}{\fn, \likeparam} = \Normal(\yn; \Wn \gn, \mat{\Sigma}_y)$ where $\W_{(n)ij}=w_{ij} \quad \forall n=1,\dots, \n$ and  $\Qg=\p$. Kernels for all models are presented at Table \ref{tab:modkerns}. We maintain similar kernel specification across models. All kernels are based on the specification described at \S \ref{sec:ggp-solar}.  
\begin{table}[t]
	\centering
	\caption{Latent function kernel specifications for \ggp and benchmark models. 
		$\kernel_{Per.}(t,s)$ is a periodic kernel applied to a time index; 
		$\kernel_{RBF}(\lags_{it}, \lags_{is})$ is a radial basis function kernel applied to recent lagged power output;
		$\kernel_{RBF-Ep.(2)}(\h_i, \h_j)$ is a separable, multiplicative radial basis-Epanechnikov function kernel applied to cross-site spatial features (latitude and longitude).
		}
\noindent \begin{tabular}{@{}ll@{}}
	\toprule
	model & $\kj$ (benchmark models)\\
	\midrule
	\igp &$\kernel_{Per.}(t,s)\kernel_{RBF}(\lags_{it}, \lags_{is})$\\
	\mtg &$ \kernel_{Per.}(t,s) \kernel_{RBF}(\lags_{it}, \lags_{js}) \kernel_{RBF-Ep.(2)}(\h_i, \h_j)$\\
	\lcm &$\kernel_{Per.}(t,s)\kernel_{RBF}(\lags_{it}, \lags_{is}), \quad i=1\text{ to }\p$.\\
	\gprn & \\
	$\W_{i, j}$&$\kernel_{Per.}(t,s)\kernel_{RBF}(\lags_{it},\text{ }\lags_{is})$ \\
	$\gj$ & $\kernel_{RBF}(\lags_{it}, \lags_{is}),\quad  i=1\text{ to }\p$ \\
	& \\
	\ggp & $\krxh$ (\ggp solar)\\
	\midrule
	$\W_{i,:}$ &$\kernel_{Per.}(t,s)\kernel_{RBF}(\lags_{it},\text{ }\lags_{is})\kernel_{RBF-Ep.(2)}(\h_i, \h_j)$ \\
	$\gj$ &$\kernel_{RBF}(\lags_{it}, \lags_{is}),\quad  i=1\text{ to }\p$ \\
	& \\
	\ggp & $\krxh$ (\ggp wind)\\
	\midrule
	$\W_{i, j \neq i}$ &$\kernel_{Per.}(t,s)\kernel_{RBF}(\lags_{it},\text{ }\lags_{is})\kernel_{RBF-Ep.(2)}(\h_i, \h_j)$ \\
	$\W_{i, i}$ &$\kernel_{Per.}(t,s)\kernel_{RBF}(\lags_{it},\text{ }\lags_{is})$ \\
	$\gj$ &$\kernel_{RBF}(\lags_{it}, \lags_{is}),\quad  i=1\text{ to }\p$ \\
	\bottomrule
\end{tabular}
\label{tab:modkerns}%
\end{table}%
Models are presented for diagonal and full \mog posterior specifications, with $\k=1$.
In the case of the \ggp, to maintain the scalable specification, we adopt a Kronecker construction of the full posterior for each group $\bR$ in line with the prior specification.

To compare model performance under equivalent settings, we consider the complexity of the different approaches and standardize model settings by reference to a consistent target computational complexity per iteration. In our variational  framework, the time complexity is dominated by algebraic operations with cubic complexity on the number of inducing inputs $\m$. We therefore set $\q\m^3 = \bR\m^3  = 20 \times (200)^3$ for \adelautumn and \adelsummer models, $\q\m^3 = \bR\m^3  = 24 \times (200)^3$ for \sydautumn, and adjust the number of inducing points, $\m$, accordingly (Table \ref{tab:modconsts}).
\begin{table}[t]
	\centering
	\caption{Key Constants for \ggp and Benchmark Models. Output dimension ($\p$); number of node functions ($\Qg$, \lcm, \gprn and \ggp only); number of latent functions ($\q$);
		 number of latent function groups ($\bR$); dimension of inducing point matrices $\Krzz$ ($\m$); and total inducing points (Agg. Ind.) $\m$ has been set to obtain roughly the same computational cost per iteration across all models.}
	\begin{tabular}{@{}lrrrrrr@{}}
		\toprule
		Model & $\p$ & $\Qg$ & $\q$ & $\bR$ &$\m$ &Agg. Ind. \\
		\midrule
		 \multicolumn{7}{l}{\adelautumn, \adelsummer}\\
			\igp & 1 & 1& & 1& 543 & 543\\
			\mtg & 1& 1& & 1& 543 & 543\\
			\lcm &10 &10 & 10 & 10 &252 & 2,520\\
			\gprn & 10 & 10& 110 & 110& 113 & 12,463\\
			\ggp & 10 &10 & 110 & 20 & 200 & 4,000\\
			& & & & &  & \\
		 \multicolumn{7}{l}{\sydautumn}\\
				\igp & 1 & 1& & 1& 577 & 577\\
				\mtg & 1& 1& & 1& 577 & 577\\
				\lcm &12 &12 & 12 & 12 &252 & 3,024\\
				\gprn & 12 & 12& 156 & 156& 107 & 16,718\\
				\ggp & 12 &12 & 156 & 24 & 200 & 4,800\\
		\multicolumn{7}{l}{\melws}\\
		\igp & 1 & 1& & 1& 524 & 524\\
		\mtg & 1& 1& & 1& 524 & 524\\
		\lcm &6 &6 & 6 & 6 &288 & 1,730\\
		\gprn & 6 & 6& 42 & 156& 107 & 6,300\\
		\ggp & 6 &6 & 42 & 18 & 200 & 3,600\\
		\bottomrule
	\end{tabular}
	\label{tab:modconsts}%
\end{table}%

\subsubsection{Experiment settings and performance measures}
All models are estimated based on the variational framework explained in \S \ref{sec:inference}.
We optimize the \elbotext iteratively until its relative change  over successive epochs is less than $10^{-5}$ up to a maximum of 200 epochs. Optimization is performed using  \adam  \citep{kingma-et-al-2014}
with settings $\{LR = 0.005; \beta_{1}=0.09; \beta_{2}=0.99 \}$.  All data except time index features are normalized prior to optimization.
Reported forecast accuracy measures are root mean squared error (\rmse) and negative log predictive density (\nlpd). The non-Gaussian likelihood of \gprn models makes the usual analytical expression for \nlpd intractable. We therefore estimate it using Monte Carlo:
\begin{align}
\text{\nlpd} = \expectation{\qfstar}{\ln \pcond{\ystar}{\fstar}} \approx \frac{1}{\s} \sum_{s=1}^{\s} \ln \Normal(\ystar; \W_{\star}^{s}\vecg_{\star}^{s}, \likeparam) \nonumber \text{,}
\end{align}
where $\W_{\star}^{s}\vecg_{\star}^{s}$ are draws from their corresponding posterior over $\fstar$.
In addition, we compute average ranking (\mrank) over both accuracy measures (\rmse and \nlpd), and mean forecast variance (\fvar), which is critical to the use of short term forecasts as inputs to system or market management algorithms.
\subsubsection{Results}
Results for solar models are presented at Table \ref{tab:resultswide} with diagonal and full-Gaussian posterior specifications. \ggp maintains or improves point accuracy when compared to best performing benchmarks on both \rmse and \nlpd individually. For \rmse, accuracy under \ggp differs by less than one percent relative to \gprn, and similarly matches or improves on \nlpd relative to \lcm and other benchmarks. \ggp performs strongly in terms of overall accuracy across both measures, consistently achieving the highest average rank across both measures (\mrank). In contrast, competing baselines either perform well on \rmse at the expense of poor performance under \nlpd or \emph{vice versa}.
%
\begin{table}[htbp]
	\renewcommand{\tabcolsep}{3pt}
	\small
  \centering
  \caption{Forecast accuracy and variance of \ggp and benchmark models
  	using diagonal (D) and full (F) Gaussian posteriors. \mrank is mean of \rmse and \nlpd ranks and \fvar is mean forecast variance. Lower values indicate better performance for all measures.
  	* indicates significant difference from best \ggp model ((D) or (F)) based on 95 percent credible interval.}
    \begin{tabular}{lrp{0.1cm}rp{0.1cm}rrp{0.1cm}p{0.5cm}rp{0.1cm}rp{0.1cm}rrp{0.1cm}}
    \toprule
     & \multicolumn{7}{c}{\adelautumn}                           &       & \multicolumn{7}{c}{\adelsummer} \\
          & \rmse  &       & \nlpd  &       & \mrank& \fvar &       &       & \rmse  &       & \nlpd  &       & \mrank & \fvar &  \\
    \midrule
    \ggp (D) & 0.282 &       & 0.243 &       & \textbf{2.5} & 0.140 &       &       & 0.318 &       & 0.323 &       & \textbf{2.5} & 0.118 &  \\
    \ggp (F) & 0.288 & *     & 0.265 & *     & 4.0   & \textbf{0.136} & \textbf{*} &       & 0.321 & *     & 0.352 & *     & 4.0   & \textbf{0.113} & \textbf{*} \\
   \lcm (D) & 0.294 & *     & 0.240 &       & 4.0   & 0.162 & *     &       & 0.325 & *     & 0.332 & *     & 4.5   & 0.165 & * \\
   \lcm(F) & 0.293 & *     & \textbf{0.240} & \textbf{*} & 3.0   & 0.160 & *     &    & 0.323 & *     & \textbf{0.323} &  & 3.0   & 0.158 & *    \\
    \gprn (D) & \textbf{0.278} & \textbf{*} & 0.311 & *     & 3.0   & 0.173 & *     &       & \textbf{0.315} & \textbf{*} & 0.376 & *     & 3.0   & 0.152 & * \\
    \gprn (F) & 0.283 &       & 0.320 & *     & 4.5   & 0.174 & *     &       & 0.316 & *     & 0.382 & *     & 4.0   & 0.152 & * \\
    \mtg (D) & 0.301 & *     & 0.337 & *     & 7.0   & 0.174 & *     &       & 0.444 & *     & 0.675 & *     & 10.0  & 0.256 & * \\
    \mtg (F) & 0.304 & *     & 0.376 & *     & 9.0   & 0.206 & *     &       & 0.441 & *     & 0.674 & *     & 9.0   & 0.267 & * \\
    \igp (D) & 0.315 & *     & 0.368 & *     & 9.0   & 0.177 & *     &       & 0.341 & *     & 0.415 & *     & 7.5   & 0.153 & * \\
    \igp (F) & 0.314 & *     & 0.370 & *     & 9.0   & 0.183 & *     &       & 0.343 & *     & 0.414 & *     & 7.5   & 0.156 & * \\
          &       &       &       &       &       &       &       &       &       &       &       &       &       &       &  \\
    \midrule
     & \multicolumn{7}{c}{\sydautumn}                                    &       & \multicolumn{7}{c}{\melws} \\
          & \rmse  &       & \nlpd  &       & \mrank & \fvar &       &       & \rmse  &       & \nlpd  &       & \mrank & \fvar &  \\
    \midrule
    \ggp (D) & 0.284 &       & \textbf{0.257} &       & \textbf{2.5} & 0.157 &       &       & 0.454 & *     & 0.670 & *     & 2.5   & 0.282 & * \\
    \ggp (F) & 0.298 & *     & 0.286 & *     & 6.0   & \textbf{0.142} & *     &       & \textbf{0.450} &       & \textbf{0.661} &       & \textbf{1.0} & \textbf{0.281} &  \\
    \lcm (D) & 0.310 & *     & 0.273 & *     & 6.5   & 0.180 & *     &       & 0.465 & *     & 0.675 & *     & 4.0   & 0.305 & * \\
    \lcm(F) & 0.302 & *     & 0.257 &       & 5.5   & 0.178 & *     &       & 0.465 & *     & 0.677 & *     & 5.0   & 0.308 & * \\
    \gprn (D) & 0.281 & *     & 0.323 & *     & 3.5   & 0.185 & *     &       & 0.460 & *     & 0.702 & *     & 3.5   & 0.308 & * \\
    \gprn (F) & 0.284 &       & 0.326 & *     & 5.5   & 0.187 & *     &       & 0.455 & *     & 0.693 & *     & 5.0   & 0.306 & * \\
    \mtg (D) & \textbf{0.280} & & 0.342 & *     & 5.0   & 0.207 & *     &       & 0.474 & *     & 0.735 & *     & 10.0  & 0.348 & * \\
    \mtg (F) & 0.283 &       & 0.360 & *     & 6.5   & 0.219 & *     &       & 0.472 & *     & 0.728 & *     & 8.0   & 0.336 & * \\
    \igp (D) & 0.286 &       & 0.340 & *     & 7.5   & 0.204 & *     &       & 0.472 & *     & 0.721 & *     & 7.0   & 0.335 & * \\
    \igp (F) & 0.286 &       & 0.335 & *     & 6.5   & 0.202 & *     &       & 0.473 & *     & 0.728 & *     & 9.0   & 0.346 & * \\
    \bottomrule
    \end{tabular}%
  \label{tab:resultswide}%
\end{table}%

The benefit of regularization under the \ggp is clear when considering mean forecast variance, which is lower under \ggp than all benchmark models for all experiments. Compared to the un-grouped \gprn (\lcm), variance for solar forecasts is reduced by 18 to 24 (13 to 40) percent under the most accurate \ggp model. 

We test statistical significance of differences in performance discussed above via 95 percent intervals estimated by Monte Carlo.
\footnote{The Monte Carlo procedure tests for significance of differences between model performance metrics by repeatedly resampling from the test data and recalculating the difference in performance metrics across models for each sample (sample size is $\n_{test}$). Differences are deemed statistically significantly different from zero where the null value falls outside the interval defined by percentiles (0.025, 0.975) of the constructed empirical distribution.} 
Results of the analysis show that \rmse under \gprn is statistically significantly lower than under \ggp for solar datasets. In all other cases, \rmse is either not significantly different or significantly higher than under \ggp. Results are similar for \nlpd, which is statistically significantly lower under \lcm for two of three datasets, and otherwise higher or not significantly different. 

\begin{figure}[htbp]
	\centering
	\includegraphics[width=0.5\linewidth]{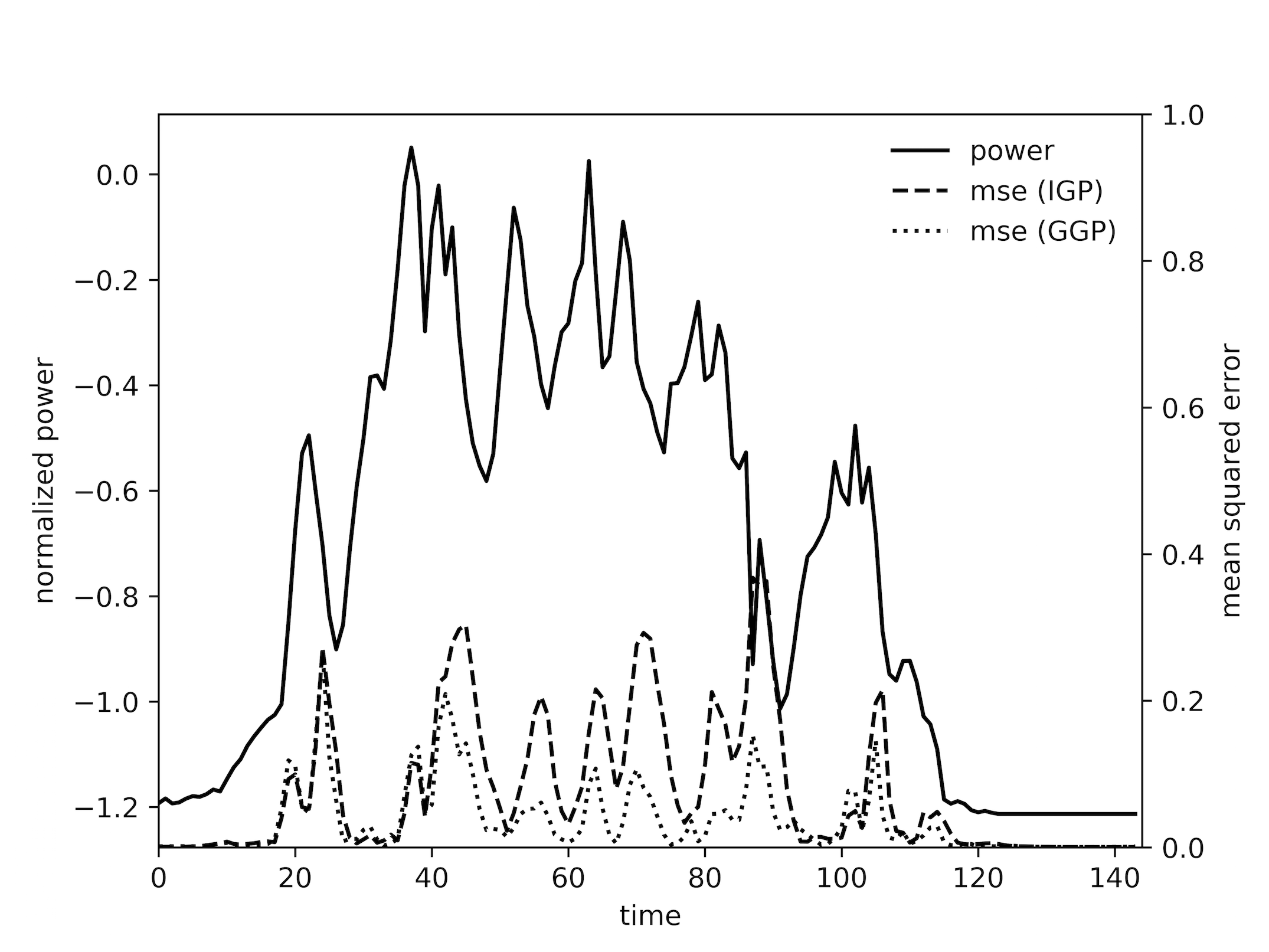}
	\caption{Mean squared error under the \ggp and \igp approaches for \adelautumn. Results shown for a single day with variable cloud cover causing high variability in power output.}
	\label{fig:ggp_igp}
\end{figure}

With the exception of the \mtg model, all multi-task models consistently improve on the naive independent forecast models. Figure \ref{fig:ggp_igp} illustrates the benefit observed under the \ggp (and other multi-task models) in reducing large forecast errors associated with variable weather conditions.

\subsection{Wind speed forecasting}
Wind variability shares characteristics with solar variability, as discussed in \S \ref{sec:intro}, with similar approaches applied to the problem of short term forecasting \citep{Widen-et-al-solar-2015}. 
We test our \ggp method forecasting ground wind speed thirty minutes ahead at six weather stations in Victoria, Australia, within an approximately 30 by 40 kilometer area. Data are half-hourly wind speed readings collected over an eight month period. The \melws data present an interesting challenge, with frequent missing and noisy observations (Figure \ref{fig:data_example}). After filtering, we have 4000 training points and 1024 test points per station.

We adopt the same kernel and feature definitions as for solar (Table \ref{tab:modkerns}) however use a different grouping structure for the \ggp. We allow functions on the diagonal of $\W$ to be independent and group off-diagonal functions within each row. This structure for each task $j$ allows weight placed on its `own' univariate node function $\gj$ to be independent of weights placed over remaining sites, which are still spatially smoothed. We similarly adjust the number of inducing points, $\m$ to test models under equivalent settings, specifically setting $\q\m^3 = \bR\m^3  = 18 \times (200)^3$.

Results for \melws are presented at Table \ref{tab:resultswide} with diagonal and full-Gaussian posterior specifications. On this dataset \ggp outperforms all other models on all measures including point accuracy (\nlpd and \rmse), overall accuracy as measured by mean model ranking across both \rmse and \nlpd, and forecast variance. Consistent reductions in variance are observed for the \melws dataset, ranging from 7 to 25 percent improvements over competing models. As for solar, confidence intervals are constructed via Monte Carlo. For \melws, all differences in model performance are confirmed to be statistically significant.

\paragraph{Comparison to approach of \citet{remes-pmlr-2017}}
In addition to the above benchmarks, estimated using the generic sparse, variational inference framework, we also consider the approach of \citet{remes-pmlr-2017}. Since this method is a variational approach with complexity of $\bigO \left( (\q\n)^{3} \right)$, which does not use inducing points, in order to fit a model under equivalent complexity conditions, we take a subset of the training data such that $(\q\n)^{3}$ approximates the settings above. 
We estimate a model for the \melws dataset, which has a manageable number of tasks. We set $\q=2$. Equivalent complexity would imply $\n=66$ for \melws, however we limit the minimum data size at $\n=200$.

We utilize the model implementation made available by the authors and allow all parameters to be optimized \footnote{The Matlab implementation of the model presented in \citep{remes-pmlr-2017} is available at https://github.com/sremes/wishart-gibbs-kernel. Models presented in \citep{remes-pmlr-2017} for the \gprn likelihood were estimated using fixed parameter values for weight latent functions.} The model gradient was optimized over 50 iterations, repeated ten times using different random parameter initial values. The model with the best performance (lowest objective function) was used to generate predictions.

The estimated value for \rmse was 0.88 for the \melws dataset, significantly higher than results under the \ggp. 

\section{Timed Experiments \label{sec:timedexpts}}
To further examine the properties of the \ggp model in relation to existing scalable multi-task methods, we conduct a series of timed experiments. We re-estimate models for the same forecasting problems as presented at \S \ref{sec:expts} and, for each epoch completed, capture time and performance measures at that point. The goal of the analysis is to evaluate time taken for the \ggp approach to achieves gains in forecast accuracy relative to the independent \gprn and other benchmarks, as well as final forecast performance attained upon completion.

We reiterate that, as for experiments presented at \S \ref{sec:expts}, the number of inducing points for each model is set to approximately standardize computational complexity per iteration.

All models are estimated on a multi-GPU machine with four NVIDIA TITAN Xp graphics cards (memory per card 12GB; clock rate 1.58 GHz).  Experiments were run until either convergence criteria were reached (see \S \ref{sec:expts}), or to a maximum of 500 epochs or 300 minutes runtime (these constraints were set conservatively based on previous experimental results). Starting values for common components were set to be equal across all models. Optimization settings were as for all other experiments.

\subsection{Results of timed experiments}
\begin{figure}[htbp]
	\centering
	\includegraphics[width=0.9\linewidth]{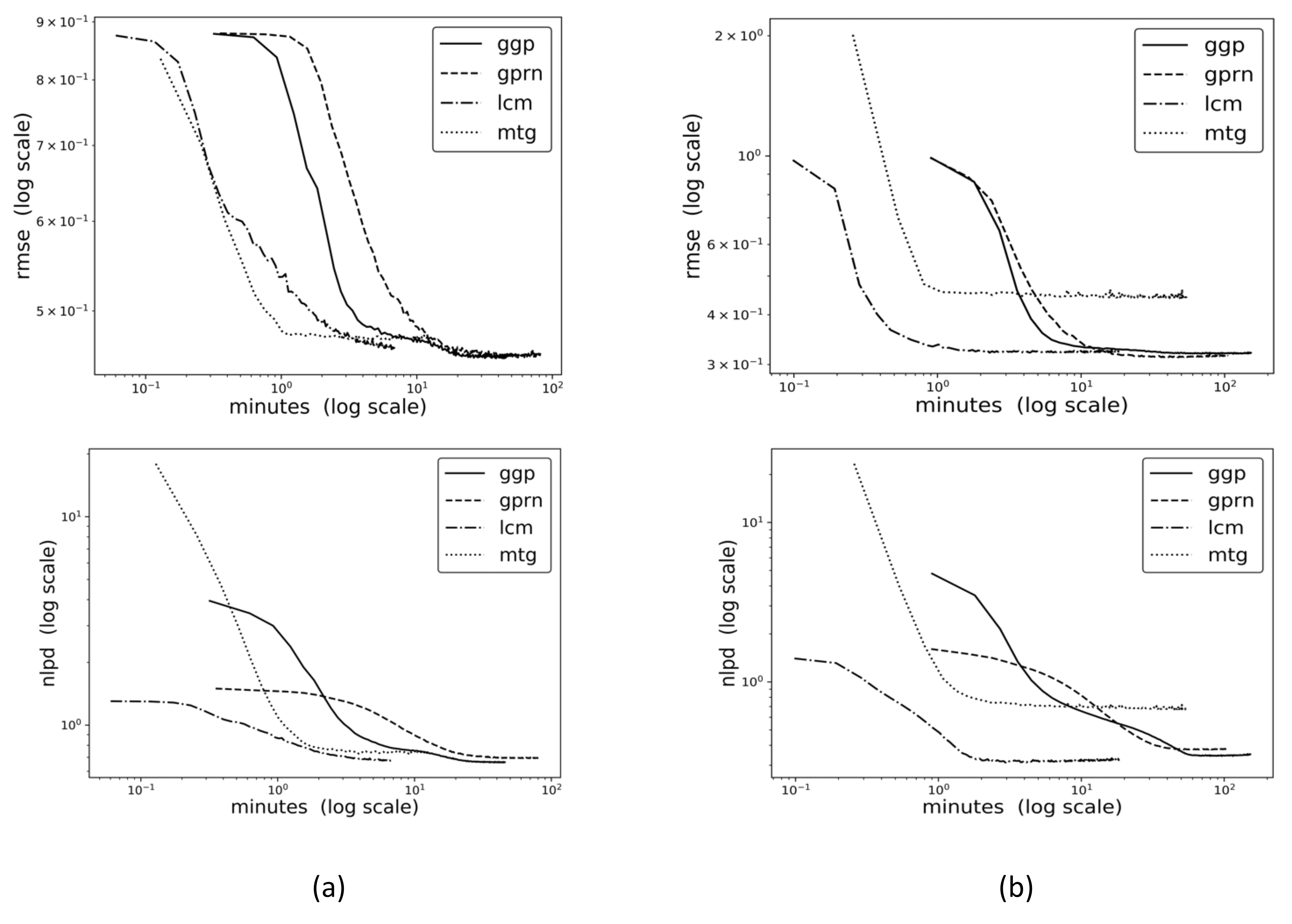}
	\caption{Forecast accuracy for \melws (a) and \adelsummer (b) datasets over optimization time in minutes for all multi-task benchmark models. \rmse and \nlpd are recorded after each epoch during optimization.}
	\label{fig:timedresults}
\end{figure}

Representative rates of improvement in performance measures over time are presented at Figure \ref{fig:timedresults} for two datasets, \melws and \adelsummer. These datasets were selected since results for \adelautumn and \sydautumn were similar to those for \melws. Results are shown for all multi-task benchmark approaches with full variational posterior specifications (similar results were obtained for the diagonal posterior setting). Performance metric values shown are recorded at the end of each epoch (hence the first value for each model is recorded at different times, being the time taken to estimate the initial epoch) and adjusted for calculation time for performance capture.

For performance at a given point in time, results suggest a consistent ranking across models tested. We observed that \ggp achieves higher forecast accuracy significantly faster than \gprn in the majority of cases, with a few cases performing similarly to \gprn. Specifically, for all datasets except \adelsummer, \rmse reduces significantly faster under the \ggp method relative to the \gprn, and \nlpd for the \ggp surpasses \gprn relatively early in the optimization. Relative rates of improvement in \rmse and \nlpd as shown for \melws at Figure \ref{fig:timedresults} provide a typical example of the performance difference between the two models. 

In terms of final accuracy, over the four datasets considered, results confirm the general rankings of final model accuracy as shown in Table \ref{tab:resultswide}. Specifically, \gprn achieves the best accuracy in terms of \rmse in two of four cases (\adelautumn and \adelsummer), while \ggp improves on \gprn in two cases (\sydautumn and \melws, with \mtg performing best on \sydautumn).  Considering both speed and final accuracy together, \ggp dominates the \gprn, achieving lower \rmse and \nlpd in a significantly shorter time than \gprn in the majority of cases. In some cases, \gprn after some time will overtake \ggp to achieve a slightly better result on \rmse, however in no case achieves a better result on \nlpd. The results of timed experiments are therefore consistent with an improvement over the \gprn in terms of speed of convergence without loss of accuracy in terms of \nlpd, and minor loss of accuracy in terms of \rmse.

With respect to the \lcm and \mtg models, these methods achieve improvements in forecast accuracy significantly more quickly than the \ggp and \gprn. Consistent with results shown at Table \ref{tab:resultswide}, \lcm achieves lower or similar \nlpd to the \ggp, with \ggp outperforming \lcm in two of four cases (\sydautumn and \melws). However, we note that as show in Figure \ref{fig:timedresults}, the \lcm converges relatively prematurely, and never achieves \ggp or \gprn performance on \rmse. A similar phenomenon was observed to a greater degree for the \mtg, which converges quickly but achieves poor accuracy relative to other models on both \rmse and \nlpd, the exception being \rmse for the \sydautumn dataset.

Across the datasets considered, the \ggp approach tends to achieve better forecast accuracy than the \lcm where data are noisier, consistent with improved accuracy where the data require a more expressive model than the (fixed-weight) \lcm approach. For example, the \adelsummer dataset has significantly less noise relative to Autumn and wind datasets (Figure \ref{fig:data_example}). Consequently, \gprn, \lcm and \ggp all perform similarly for the \adelsummer dataset in terms of final accuracy, but \lcm is significantly faster, suggesting there is little advantage from a more costly, expressive model such as \gprn or \ggp.  In contrast, for noisier datasets, \ggp and \gprn continue to improve over \lcm, and \ggp does so at a faster rate than the \gprn. Figure \ref{fig:timedresults} illustrates the typical relative accuracy over time of multitask models.

\section{Discussion}
\label{sec:discussion}
We have proposed a general multi-task \gptext model, where groups of functions are coupled \emph{a priori}. Our approach  allows for input-varying covariance across tasks governed by kernels and features and, by building upon sparse variational methods and exploiting Kronecker structures, our inference method is inherently scalable to a large number of observations.

We have shown the applicability of our approach to forecasting short term distributed solar power and wind speed at multiple locations, where it matches or improves point forecast performance of single-task learning approaches and other multi-task baselines under similar computational constraints while improving quantification of predictive variance. We have also demonstrated that our approach can yield important reductions in time taken to achieve the same accuracy relative to the equivalent model without coupled priors. In general, the \ggp strikes a balance between flexible, task-specific parameterization and effective regularization via structure imposed in the prior. 

While we focus on \apriori spatial dependencies, we emphasize that other grouping structures and kernels, likelihood functions or applications are possible. For example, non-spatial covariates in other domains, or grouping of functions according to clusters of tasks, could be adopted.
\subsubsection*{Acknowledgments}

This research was conducted with support from the Cooperative Research Centre for Low-Carbon Living in collaboration with the University of New South Wales and Solar Analytics Pty Ltd.

\bibliographystyle{apalike}
\bibliography{references}

\end{document}